\documentclass[letterpaper, 10 pt, conference]{ieeeconf}  

\IEEEoverridecommandlockouts                              

\usepackage{tabularx}
\usepackage{textcomp}
\usepackage{stfloats}
\usepackage{url}
\usepackage{verbatim}
\usepackage{graphicx}
\usepackage{cite}
\usepackage{booktabs}
\usepackage{tabularx}
\usepackage{multirow}
\usepackage{tikz}
\usepackage{makecell}
\usepackage{rotating}
\usepackage{amsmath}
\usepackage{xcolor}
\usepackage{soul}
\usepackage{threeparttable}
\usepackage{tabu}                     
\usepackage{multirow}                 
\usepackage{multicol}                 
\usepackage{multirow}                
\usepackage{float}                    
\usepackage{makecell}                 
\usepackage{booktabs}                 
\usepackage{wasysym} 

\usepackage{siunitx}
\definecolor{gray}{rgb}{0.5, 0.5, 0.5} 
\usepackage[colorlinks, linkcolor=blue, anchorcolor=blue, citecolor=blue]{hyperref}
\title{\LARGE \bf
Leveraging Geometric Prior Uncertainty and Complementary Constraints for High-Fidelity Neural Indoor Surface Reconstruction
}

\author{
Qiyu Feng$^{1*}$, Jiwei Shan$^{2*}$, Shing Shin Cheng$^{2}$ and Hesheng Wang$^{1}$
\thanks{*The first two authors contributed equally. This work was supported in part by the Natural Science Foundation of China under Grant 62361166632, 62225309, 62073222, and U21A20480, in part by Innovation and Technology Commission of Hong Kong (ITS/235/22) and in part by Multi-scale Medical Robotics Center, InnoHK. Corresponding Authors: Hesheng Wang, Shing Shin Cheng.}
\thanks{$^{1}$ Department of Automation, Shanghai Jiao Tong University, Shanghai 200240, China.}%
\thanks{$^{2}$ Department of Mechanical and Automation Engineering and T Stone Robotics Institute, The Chinese University of Hong Kong, Hong Kong.}%
}

\begin{document}

\maketitle
\thispagestyle{empty}
\pagestyle{empty}

\begin{abstract}

Neural implicit surface reconstruction with signed distance function has made significant progress, but recovering fine details such as thin structures and complex geometries remains challenging due to unreliable or noisy geometric priors. Existing approaches rely on implicit uncertainty that arises during optimization to filter these priors, which is indirect and inefficient, and masking supervision in high-uncertainty regions further leads to under-constrained optimization. To address these issues, we propose GPU-SDF, a neural implicit framework for indoor surface reconstruction that leverages geometric prior uncertainty and complementary constraints. We introduce a self-supervised module that explicitly estimates prior uncertainty without auxiliary networks. Based on this estimation, we design an uncertainty-guided loss that modulates prior influence rather than discarding it, thereby retaining weak but informative cues. To address regions with high prior uncertainty, GPU-SDF further incorporates two complementary constraints: an edge distance field that strengthens boundary supervision and a multi-view consistency regularization that enforces geometric coherence. Extensive experiments confirm that GPU-SDF improves the reconstruction of fine details and serves as a plug-and-play enhancement for existing frameworks. 
Source code will be available at https://github.com/IRMVLab/GPU-SDF

\end{abstract}

\section{INTRODUCTION}

Three-dimensional surface reconstruction from multi-view images is a long-standing challenge in computer vision and graphics. Accurate and dense geometry is crucial for applications such as AR/VR systems, robotic navigation and embodied intelligence. 
Traditional methods have made significant progress but still suffer from incomplete reconstructions and difficulty with textureless  surfaces. 
Recently, differentiable rendering approaches—most notably Neural Radiance Fields \cite{nerf} and 3D Gaussian Splatting \cite{3DGS}—have enabled photorealistic view synthesis, yet they often struggle to recover explicit surface geometry. To overcome this, researchers have explored neural signed distance functions (Neural SDF) \cite{SDF}, which represent surfaces implicitly as MLP-based signed distance fields optimized via differentiable rendering, enabling higher-fidelity reconstruction. Recent methods \cite{monosdf,OCC-SDF,neuris,DebSDF,NDSDF,dtc_nerf1,dtc_nerf2} further improve results in textureless regions and complex geometries by incorporating monocular geometric predictions as additional priors. Despite these advances, key challenges remain.

\begin{figure}[!t] 
\centerline{\includegraphics[width=1\linewidth]{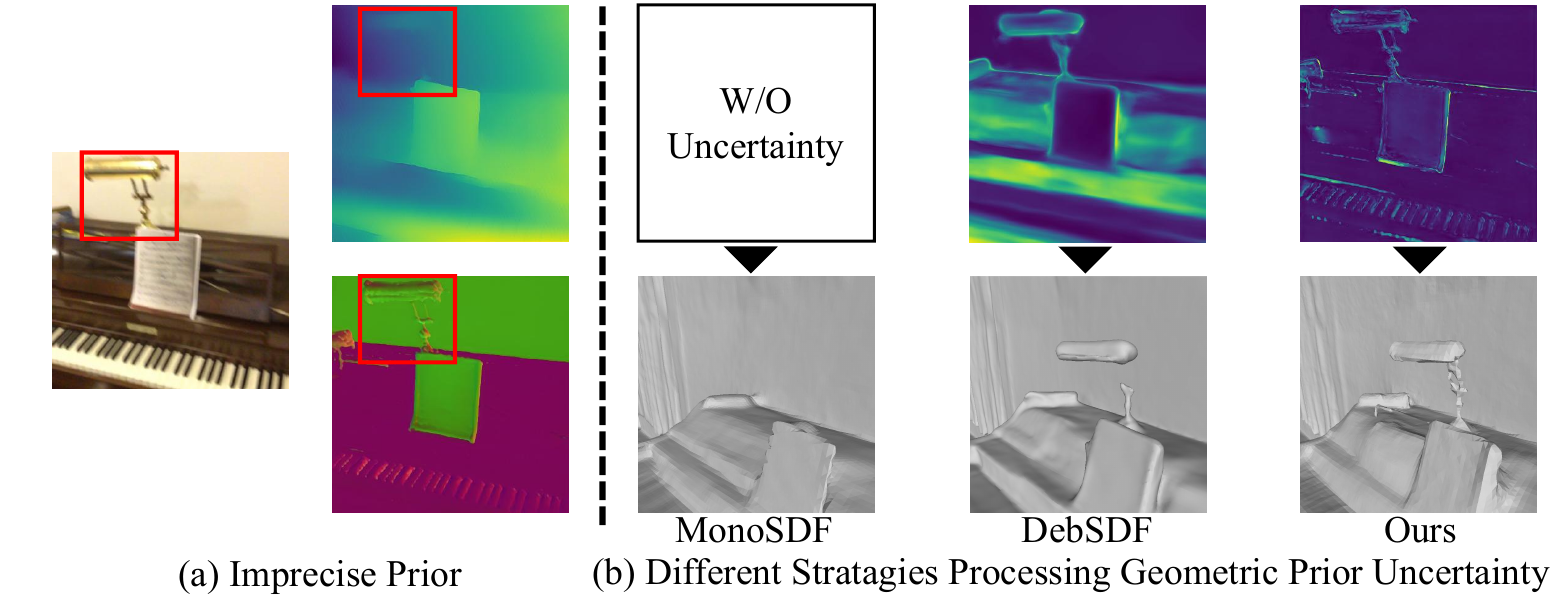}}
\vspace{-5pt}
\caption{(a) Monocular geometric priors for 3D reconstruction are often imprecise, especially for thin structures.  
(b) Comparison of different strategies to process geometric prior uncertainty. MonoSDF \cite{monosdf} uses priors directly, without handling uncertainty. DebSDF \cite{DebSDF} relies on an implicit uncertainty that emerges during optimization, discarding supervision in unreliable regions and forcing reliance solely on RGB cues. In contrast, our method explicitly estimates prior uncertainty at the outset. We then employ an uncertainty-guided loss to modulate the influence of priors according to their reliability, rather than discarding them. Together with the additional geometric constraints we introduce, our method reconstructs fine-grained structures more effectively.}

\label{fig:intro} 
\vspace{-15pt}
\end{figure}

As shown in Fig.~\ref{fig:intro}, existing Neural SDF methods can effectively reconstruct the overall indoor structure, but they still struggle to recover fine details such as chair legs and railings. 
Prior studies (e.g., DebSDF~\cite{DebSDF}) attribute these failures to several factors, including errors in monocular geometric priors due to domain gaps, the lack of multi-view consistency in independently predicted priors, and the low sampling probability of thin structures. To alleviate these issues, DebSDF proposes a strategy of estimating uncertainty during the SDF optimization process. Specifically, they use the SDF model's own emergent uncertainty about the geometry to filter unreliable supervision or reweight ray sampling. While this strategy improves robustness, it suffers from a critical limitation:
\textbf{(1) Ignoring Explicit Prior Uncertainty:} This filtering process relies solely on an \textit{implicit} uncertainty that emerges \textit{during} the SDF optimization. This uncertainty reflects the model's own difficulty in fitting the geometric priors, rather than being a direct, upfront assessment of the priors' inherent quality. Consequently, the model must first "learn" to be uncertain by struggling with noisy or inconsistent data—a process that is both indirect and inefficient. This conflation of the model's learning state with the prior's intrinsic quality can lead to suboptimal decisions: confidently incorrect priors might be retained if they are internally consistent, while weakly informative but correct priors might be discarded prematurely. 
\textbf{(2) Under-constrained optimization in high-uncertainty regions:} When geometric supervision is masked out, the learning signal for recovering scene structure becomes primarily reliant on RGB. This is often insufficient for recovering accurate geometry in textureless or thin structures where RGB cues are weak or ambiguous.

To overcome these challenges, we propose \textbf{GPU-SDF}, a neural implicit reconstruction method with two key innovations for high-quality indoor 3D reconstruction.
First, to reduce reliance on indirect, model-derived uncertainty, we introduce a self-supervised approach that explicitly estimates the confidence of geometric priors at the outset, without external networks. This separates prior quality assessment from the learning state of the SDF model. With this explicit confidence, we design a new {uncertainty-guided geometric consistency loss}. Instead of discarding noisy supervision, this loss adjusts the influence of priors based on their estimated reliability. In this way, the model still learns from weak but useful signals and mitigates degradation caused by missing supervision.
Second, to address under-constrained optimization in high-uncertainty regions, we add two complementary geometric constraints. An {edge distance field} provides robust scene edge information as an auxiliary cue. In parallel, a {multi-view consistency regularization} enforces coherent geometry across different viewpoints. Together, these constraints supply the guidance needed to reconstruct thin and fine structures where RGB cues alone are ambiguous or unreliable.
Moreover, our framework is modular and can be seamlessly integrated as a plug-in into existing neural SDF reconstruction pipelines, thereby enhancing their reconstruction performance. The main contributions are as follows:
\begin{itemize}
    \item We propose a novel self-supervised uncertainty estimation method combined with an uncertainty-guided geometric supervision strategy, which preserves weak geometric signals in high-uncertainty regions and alleviates degradation caused by discarding priors.  
    \item We design an edge distance field and a multi-view consistency regularization that strengthen supervision in under-constrained regions, improving the reconstruction of thin and fine structures.   
    \item We conduct extensive experiments on challenging benchmarks, demonstrating not only the effectiveness of our method but also its plug-and-play ability to enhance existing SDF-based frameworks.
\end{itemize}

\section{Related Works}

\subsection{Neural Implicit and Gaussian Splatting based Indoor Surface Reconstruction}

Recently, neural implicit methods have received significant attention for representing scene geometry. NeRF-type approaches~\cite{nerf,volsdf,Neus,neuralangelo,voxurf,unisurf} encode coordinate-based density and appearance using multilayer perceptrons (MLPs), represent scene geometry through signed distance functions (SDFs)~\cite{SDF}, and extract surface geometry via marching cubes~\cite{mcubes}.
For indoor scenes, neural implicit reconstruction methods using only monocular RGB images as input often struggle to reconstruct low-texture surfaces and complex geometries. To address this issue, additional priors are incorporated to constrain scene geometry, such as the Manhattan World assumption~\cite{manhattanSDF}, depth and normal priors~\cite{monosdf,neuris}, or semantic information~\cite{SemanticNeRF} to regularize textureless regions. Other approaches introduce alternative geometric representations or additional regularization terms~\cite{NDSDF,SRDF,OCC-SDF,UDF,neurodin,RayDF,tsdf-neuralrgbd,nicer-slam} to improve reconstruction quality. 
In parallel, reconstruction methods based on 3D Gaussian Splatting (3DGS)~\cite{3DGS} have shown strong potential for 3D reconstruction. 
Several works~\cite{2DGS,PGSR,GSRec,gaussianroom,sugar} have explored surface reconstruction within 3DGS framework, but a high-fidelity extraction technique—akin to marching cubes for NeRF's implicit fields—remains underdeveloped, resulting in suboptimal surface quality.
Hence, this paper focuses on neural implicit indoor surface reconstruction.

\subsection{Depth and Normal Uncertainty Estimation}

Monocular depth and normal estimation is a fundamental task, while it faces inherent challenges such as scale ambiguity, discontinuities, and systematic bias. Estimating the uncertainty of these predictions is therefore crucial for improving accuracy and reliability.
Existing approaches include uncertainty estimation with external networks~\cite{irondepth}, or training a network from scratch~\cite{monoprob} to predict probability distributions. However, these methods are computationally expensive. In contrast, self-supervised approaches~\cite{uncer1,uncer2} estimate uncertainty by measuring the variance between outputs generated from perturbed inputs (e.g., image flipping). Owing to their computational efficiency, this work adopts self-supervised methods for assessing the uncertainty of monocular depth and normal predictions.

\section{Methods}

\begin{figure*}[!t] 
\centerline{\includegraphics[width=\linewidth]{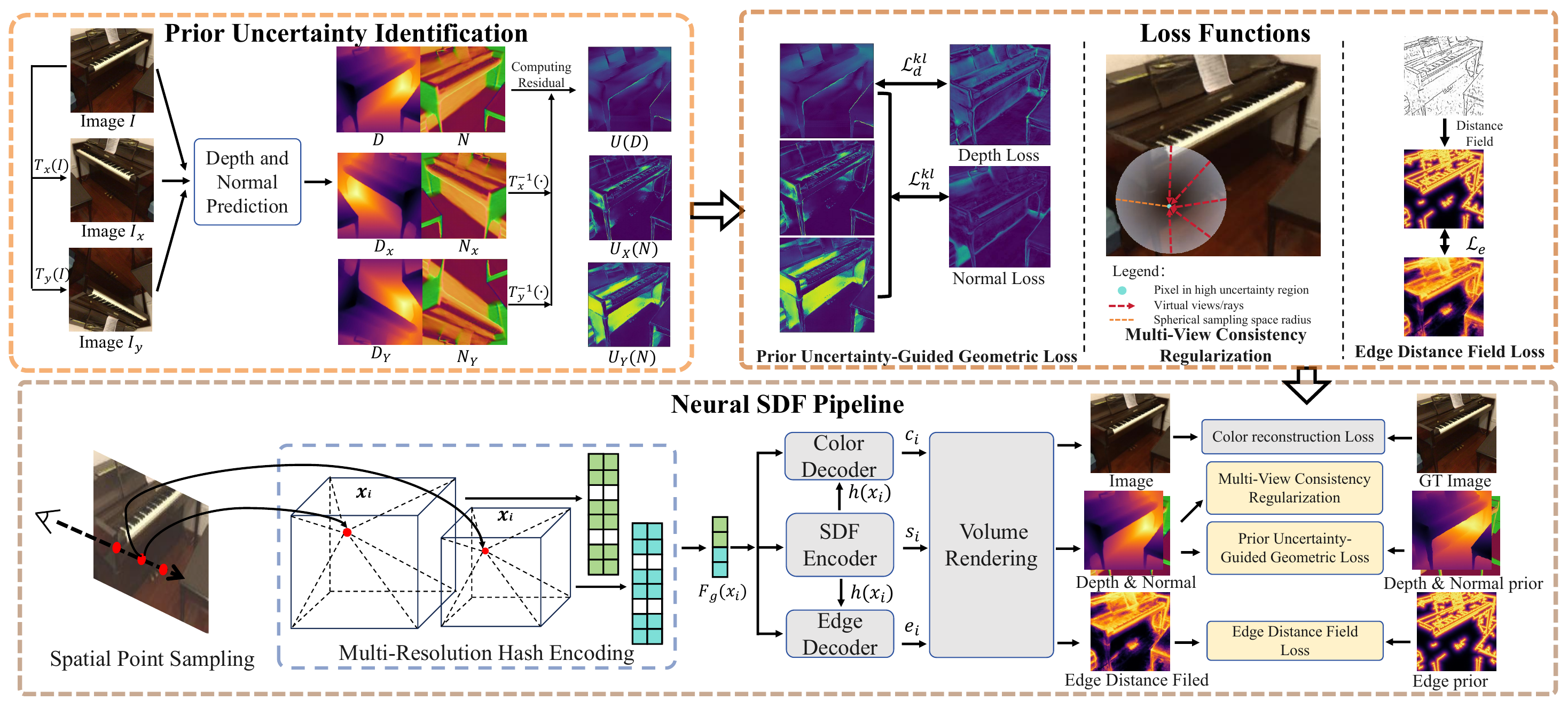}}
\vspace{-7pt}
\caption{\textbf{Overview of the GPU-SDF framework.}
GPU-SDF reconstructs high-fidelity surfaces from multi-view RGB images and initial geometric priors (e.g., depth, normals). It consists of three parts: 
\textbf{(1) Neural SDF pipeline:} builds upon existing frameworks that represent the scene with an SDF and color field, and augments them by jointly learning an edge distance field as a robust geometric cue for fine structure reconstruction. 
\textbf{(2) Prior uncertainty identification:} a self-supervised module explicitly estimates confidence of geometric priors, enabling dynamic adjustment of their influence during optimization. 
\textbf{(3) Loss functions:} an uncertainty-guided consistency loss preserves weak but useful signals, while edge distance field supervision and multi-view regularization facilitate accurate recovery of thin and fine structures.}
\label{fig:overview}
\vspace{-15pt}
\end{figure*}
Given calibrated multi-view images and depth/normal priors obtained from a pre-trained model, our objective is to develop a reconstruction framework based on Neural Signed Distance Fields (SDFs) that produces high-precision dense surfaces while preserving fine geometric details. As illustrated in Fig.~\ref{fig:overview}, the proposed GPU-SDF framework is composed of three main components: \textit{Neural SDF}, \textit{Prior Uncertainty Identification}, and \textit{Loss Functions}.  
The Neural SDF module follows the general pipeline of mainstream methods. Rather than being tied to a specific baseline, our method can be seamlessly integrated as a plugin to enhance the performance of existing neural SDF frameworks. In this work, we adopt ND-SDF~\cite{NDSDF}, one of the state-of-the-art approaches, as the base network to demonstrate the effectiveness of our design, and provide a detailed review of its key techniques in Sec.~\ref{sec:pre}. The Prior Uncertainty Identification module (Sec.~\ref{sec:unc}) introduces an effective self-supervised strategy for estimating the uncertainty of geometric priors.  
The Loss Function module (Sec.~\ref{sec:loss}) incorporates both standard losses used in neural SDF frameworks and three additional objectives specifically designed to account for prior uncertainty and enhance reconstruction, especially in regions with fine structures. 
Finally, the model optimization process is described in Sec.~\ref{sec:imp}.

\subsection{Preliminary}\label{sec:pre}
\textbf{Neural Signed Distance Fields.} Neural SDFs represent scene geometry and appearance with an implicit neural network, which is optimized through differentiable volume rendering. Specifically, given a ray \( r(t) = o + tv \) from camera origin \( o \) in direction \( v \), we sample \( N \) points \( x_i = o + t_i v \) along the ray. The implicit neural network predicts the SDF value \( s_i \) and color \( c_i \) for each point. To enable volume rendering, SDF values are converted into volume densities as~\cite{volsdf}:
\begin{equation}
\sigma(s) = 
\begin{cases} 
\frac{1}{2\beta} \exp\left(-\frac{s}{\beta}\right), & s > 0, \\
\frac{1}{\beta} - \frac{1}{2\beta} \exp\left(\frac{s}{\beta}\right), & s \leq 0,
\end{cases}
\end{equation}
where \( \beta \) is a learnable parameter.  

The rendered color, depth, and normal are then obtained through volume rendering:
\begin{equation}
\hat{C}(r) = \sum_{i=1}^{N} T_i \alpha_i c_i, \;
\hat{D}(r) = \sum_{i=1}^{N} T_i \alpha_i t_i, \;
\hat{N}(r) = \sum_{i=1}^{N} T_i \alpha_i n_i,
\label{equ:render}
\end{equation} 
where \( \alpha_i = 1 - \exp(-\sigma_i \delta_i) \) represents the opacity at \( x_i \), and $\delta_i$  denotes distance between adjacent samples. \( T_i = \prod_{j=1}^{i-1} (1 - \alpha_j) \) denotes the accumulated transmittance of light as it travels to the \( i \) point along the ray. $n_i$ is the analytical gradient of the SDF network at point \( i \). 

The network parameters are optimized by minimizing reconstruction losses with respect to input RGB images and geometric priors:
\[
\mathcal{L}_{c} = \sum_{r \in \mathcal{R}} \| \hat{C}(r) - C(r) \|_1,
\quad
\mathcal{L}_{d} = \sum_{r \in \mathcal{R}} \| w \hat{D}(r) + q - D(r) \|^2,
\]
\begin{equation}
\mathcal{L}_{n} = \sum_{r \in \mathcal{R}} \| \hat{N}(r) - N(r) \|_1 
+ \| 1 - \hat{N}(r)^\top N(r) \|_1,
\end{equation}
where \( \mathcal{R} \) represents the set of sampled rays, and \( {{C}}({r}) \) denotes the GT RGB values. \( D(r) \) and \( N(r) \) are depth and normal priors~\cite{omnidata}, and \( w, q \) are scale and shift factors estimated by least squares.  
Finally, an eikonal regularization is imposed to enforce the SDF property:
\begin{equation}
\mathcal{L}_{\text{eik}} = \frac{1}{N} \sum_{i=1}^{N} \left( \left\| \nabla s(\mathbf{x}_i) \right\|_2 - 1 \right)^2.
\end{equation}

\textbf{ND-SDF.} 
To adaptively model the reliability of normal priors across different regions of a 3D scene, ND-SDF \cite{NDSDF} extends the standard Neural SDF framework by introducing a learnable deflection field. Specifically, it predicts the deflection angle between the scene normals and the provided normal priors, represented as quaternions \(q_i\). During rendering, the quaternion at the ray--surface intersection is aggregated as $Q(r) = \sum_{i=1}^{N} T_i \alpha_i q_i$ and the rendered normal is corrected via quaternion rotation $\hat{N}^d(r) = Q(r) \otimes \hat{N}(r) \otimes Q^{-1}(r)$,
where \( \hat{N}(r) \) is the original rendered normal and \( \otimes \) denotes quaternion multiplication. The normal reconstruction loss is then built upon the deflected normal \(\hat{N}^d(r)\) and the priors:
\begin{equation}
\mathcal{L}_{n}^d = \sum_{r \in \mathcal{R}} \| \hat{N}^d(r) - N(r) \|_1 + \| 1 - \hat{N}^d(r)^\top N(r) \|_1.
\end{equation}

Furthermore, ND-SDF introduces an angle-aware reweighting mechanism for RGB, depth, and normal losses:
\begin{equation}
\mathcal{L}^{ad}_p = w_p(\Delta\theta)\,\mathcal{L}_p, \quad p \in \{c,d,n\},
\end{equation}
where \(\Delta\theta = \arccos(\hat{N}(r)\cdot \hat{N}^d(r))\) denotes the deflection angle, and \(w_p\) adjusts the confidence of each loss accordingly.

The final reconstruction objective combines the weighted color, depth, and normal terms with eikonal regularization:
\begin{equation}
\mathcal{L}_{recon} = \lambda_{c}\mathcal{L}^{ad}_{c} + \lambda_{d}\mathcal{L}^{ad}_{d} + \lambda_{n}\mathcal{L}^{ad}_{n} + \lambda_{eik}\mathcal{L}_{eik}.
\end{equation}

\subsection{Prior Uncertainty Identification}\label{sec:unc}
To explicitly evaluate the reliability of geometric priors, rather than relying on the implicit uncertainty generated during optimization in existing methods~\cite{DebSDF}, we introduce a self-supervised uncertainty estimation module. This module models uncertainty as a distribution that reflects the variance of prediction errors. Based on this explicit uncertainty, we further design an uncertainty-guided geometric consistency loss (Sec.~\ref{sec:loss}) that adaptively adjusts the influence of noisy priors, leading to more accurate and robust reconstruction.

Specifically, given an RGB image \( I \) and a pretrained depth estimation network \( D_\theta \) \cite{omnidata}, we obtain the monocular depth map as \( D = D_\theta(I) \). Existing approaches to estimate depth uncertainty usually require either retraining the model or relying on ground-truth depth~\cite{irondepth,monoprob,scratch}, both of which are impractical in many real-world scenarios. 
Therefore, we aim to design a self-supervised, post-hoc method that can be directly applied to pretrained models.  
Inspired by~\cite{uncer1,uncer2}, we introduce a simple yet effective strategy based on flip consistency. 
Let \( T_m(\cdot) \) denote a transformation along axis \( m \in \{x,y\} \), corresponding to horizontal or vertical flipping, and \( T_m^{-1}(\cdot) \) its inverse operation. Since these transformations preserve pixel-level structures, the depth prediction of the original image \( I \), denoted as \( D \), should remain consistent with the predictions of the flipped images \( I_m = T_m(I) \), denoted as \( D_m \).  To estimate depth uncertainty, we compute the depth maps for both the original and flipped images, and then realign the flipped predictions:  
\[
D = D_\theta(I), \quad 
D_x = D_\theta(I_x), \quad 
D_y = D_\theta(I_y),
\]
\begin{equation}
\tilde{D}_x = T_x^{-1}(D_x), \quad 
\tilde{D}_y = T_y^{-1}(D_y).
\end{equation}
\textcolor{black}{The pixel-wise depth uncertainty \( U(D) \in \mathbf{R}^{H \times W} \) is then defined as:
\begin{equation}
\begin{split}
U_x(D) &= {|D - \tilde{D}_x|}, U_y(D) = {|D - \tilde{D}_y|}, \\
U(D)   &= \sqrt{\frac{U_x^2(D)+U_y^2(D)}{2}} 
\end{split}
\end{equation}
Here, $U_x(D)$ and $U_y(D)$ represent the uncertainty components obtained from horizontal and vertical flips, respectively.  Then we adopt their standard deviation as a unified metric for quantifying the uncertainty of depth prior.
}
Following the same principle, we compute normal uncertainty. Given the predicted normal map \(N\), we obtain the aligned flipped predictions \(\tilde{N}_x, \tilde{N}_y\) and define:
\begin{equation}
U_x(N) = |N - \tilde{N}_x|, \quad
U_y(N) = |N - \tilde{N}_y|.
\end{equation}
Furthermore, due to the geometric relationship between depth and normals, the uncertainty along the \(z\)-axis is directly derived from the depth uncertainty:
\begin{equation}
U_z(N) = U(D).
\end{equation}

\begin{figure}[t] 
\centerline{\includegraphics[width=1\linewidth]{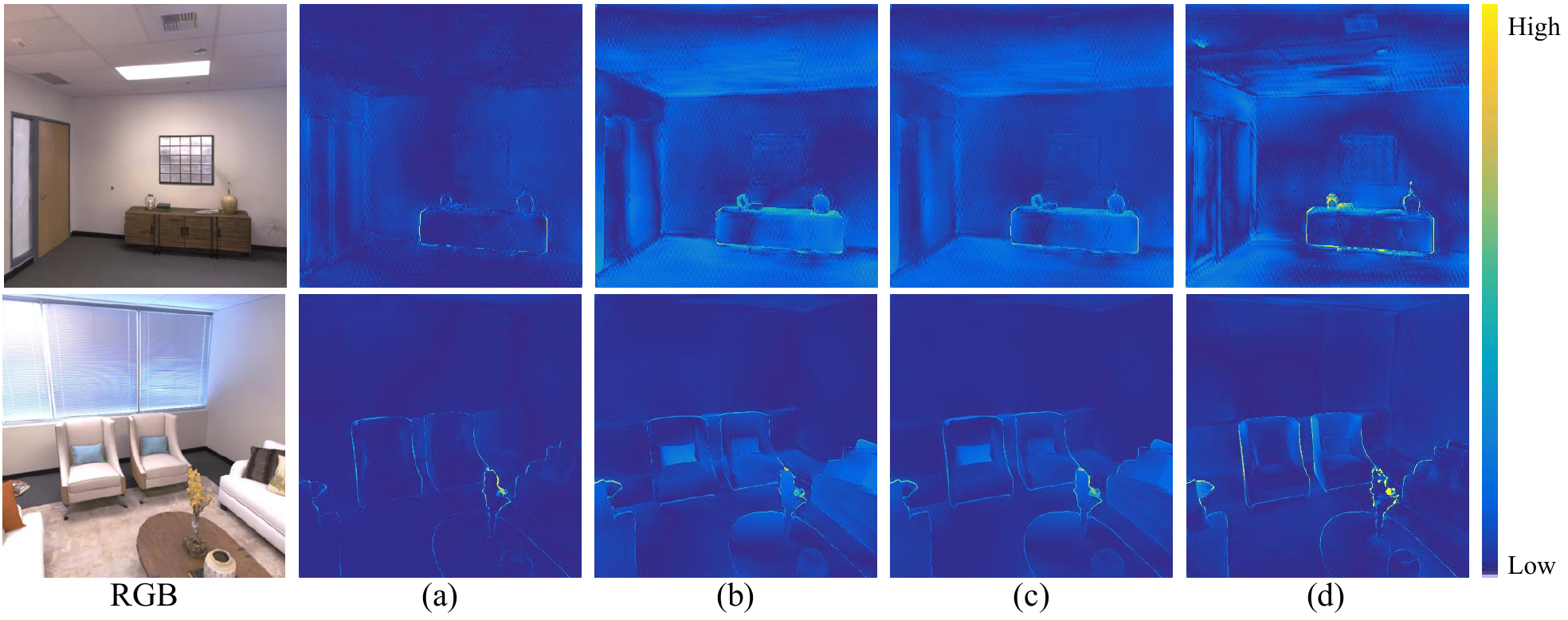}}
\vspace{-10pt}
\caption{
Visualization of our self-supervised uncertainty estimation. (a) Uncertainty from horizontal flips, $U_x(D)$; (b) uncertainty from vertical flips, $U_y(D)$; (c) the combined uncertainty, $U(D)$, which aligns closely with the ground-truth depth error map in (d).
}\label{fig:uncer}
\vspace{-20pt}
\end{figure}

Compared with prior works~\cite{uncer1,uncer2} that rely only on horizontal flips, our approach incorporates both horizontal and vertical consistency, producing a more robust uncertainty map by capturing geometric inconsistencies missed by single-axis tests (see Fig.~\ref{fig:uncer}). Furthermore, we extend this principle from depth to normal estimation, providing stronger optimization guidance and yielding higher-quality reconstructions (Sec.~\ref{sec:ab}).

\subsection{Loss Functions}\label{sec:loss}
We optimize and train our model by minimizing the loss function with respect to the implicit network parameters \( \theta \) of the neural SDF. In addition to the basic losses introduced in Sec.~\ref{sec:pre}, the estimated prior uncertainty enables two key improvements: 
1) adaptively controlling the supervision strength of prior-related losses, leading to our \textit{Prior Uncertainty-Guided Geometric Loss};  
2) introducing additional constraints in regions with high prior uncertainty, including an \textit{Edge Distance Field loss} and a \textit{multi-view consistency regularization} module.  

\textbf{Prior Uncertainty-Guided Geometric Loss.}  
Most existing methods mitigate the effect of uncertain priors by discarding supervision in high-uncertainty regions based on a predefined threshold \cite{DebSDF}. However, this simplification leads to under-constrained optimization that relies mainly on RGB supervision, often causing local blurring or structural loss in reconstruction. To overcome this limitation, we design a prior uncertainty-guided geometric loss that adaptively regularizes the contribution of geometric supervision according to its uncertainty.  Concretely, let \(R\) denote the set of sampled rays. For each ray, the prior normal and its prediction are denoted as \(N_i^d\) and \(\hat{N}_i^d\) with \(i \in \{x, y, z\}\), and the depth prior and prediction are denoted as \(D\) and \(\hat{D}\), respectively.
Inspired by the form of KL divergence, we define the following regularization losses:  
\begin{equation}
\mathcal{L}_n^{kl} = \sum_{i\in \{x,y,z\}} \sum_{r \in R} 
\left| N_i^d - \hat{N}_i^d \right| 
\cdot \ln \left( \frac{\left| N_i^d - \hat{N}_i^d \right|}{U_i(N)} \right),
\end{equation}
\begin{equation}
\mathcal{L}_d^{kl} = \sum_{r \in R} 
\left| D - \hat{D} \right| 
\cdot \ln \left( \frac{\left| D - \hat{D} \right|}{U(D)} \right).
\end{equation}
Here, $U_i(N)$ and $U(D)$ represent the prior uncertainties for normals and depth, respectively. 
This formulation dynamically scales the supervision strength: reliable priors enforce strong constraints, while uncertain priors contribute weaker but still informative regularization. 
In this way, our method avoids discarding useful information and promotes plausible geometry reconstruction even in high-uncertainty regions.

\textbf{Edge Distance Field Loss.}  
Accurately recovering sharp object boundaries and details remains a key challenge in neural surface reconstruction. This difficulty stems from the inherent unreliability of monocular geometry priors in these high-frequency regions. Without explicit guidance, the optimization can be under-constrained, leading to overly smooth or noisy geometry at object edges. To address these issues, we introduce edge information as an additional cue. Edges naturally delineate object boundaries, providing guidance for recovering fine details. They also act as complementary constraints in uncertain regions, helping to stabilize training and maintain structural fidelity.  Specifically, we employ TEED~\cite{TEED} to extract edge maps from RGB images and convert them into edge distance fields for supervision. As shown in Fig.~\ref{fig:overview}, similar to~\cite{dds-slam}, we add an edge decoder to the neural SDF framework, which shares the same architecture as the SDF decoder. For each sampled 3D point \(x_i\), the edge decoder predicts an edge value \(e_i\). These predictions are then aggregated through volumetric rendering, as in \ref{equ:render}, to compute the rendered edge distance field:  
\begin{equation}
\hat{E}(r) = \sum_{i=1}^{N} T_i \alpha_i e_i.    
\end{equation}
For supervision, we pre-compute edge distance fields from edge maps using a distance transform, which serve as pseudo ground-truth. The edge distance field loss is then defined as the L1 difference between the rendered edge distance field \(\hat{E}(r)\) and the pre-computed edge distance field \(E(r)\):
\begin{equation}
\mathcal{L}_{e} = \sum_{r \in \mathcal{R}} \| \hat{E}(r) - E(r) \|_1.
\end{equation}

\begin{figure}[t] 
\centerline{\includegraphics[width=0.7\linewidth]{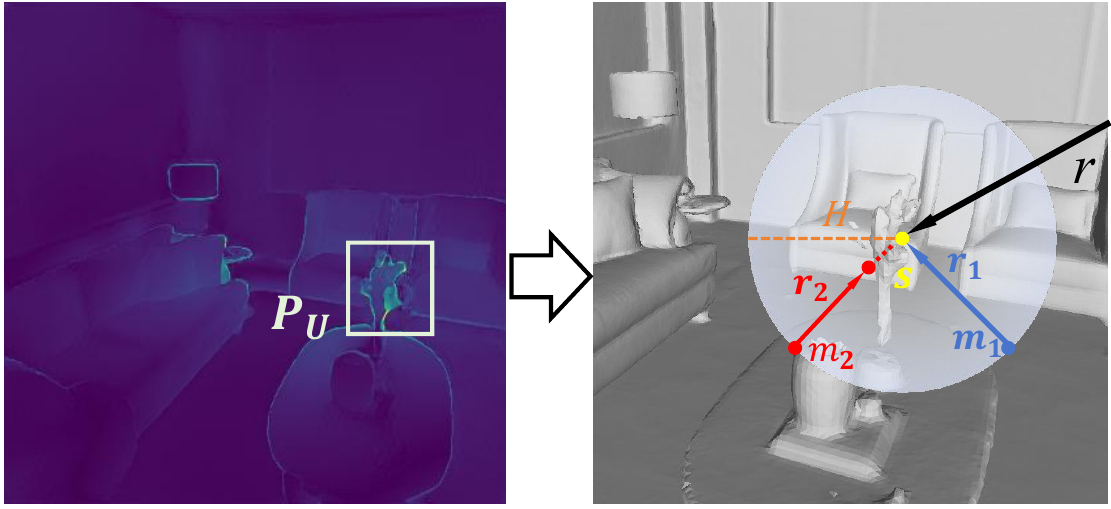}}
\vspace{-10pt}
\caption{
Illustration of the multi-view consistency regularization. For a pixel within a high-uncertainty region $P_U$, its corresponding primary ray $r$ intersects the surface at point $s$. A sphere with radius $H$ is centered at $s$. Auxiliary rays, such as $r_1$ and $r_2$, are then sampled, originating from points ($m_1$, $m_2$) on the sphere's surface. If the first surface intersection of an auxiliary ray coincides with $s$, its visibility flag is set to $v_i=1$, and it is included in the consistency loss calculation. Otherwise, the flag is set to $v_i=0$, and the ray is excluded.
}
\vspace{-17pt}
\label{fig:multi-reg}
\end{figure}
\textbf{Multi-View Consistency Regularization.}  
3D scenes exhibit an inherent property of geometric coherence: the same surface region, when observed from different viewpoints, should yield consistent depth estimates. To leverage this property, we introduce multi-view consistency regularization that activates only in high-uncertainty regions $P_U$, locally refining weak areas to improve reconstruction stability while avoiding unnecessary overhead. 

Specifically, we first identify the high-uncertainty regions $P_U$ based on the geometric prior uncertainty introduced in Sec.~\ref{sec:unc}. 
A pixel is included in $P_U$ if its depth uncertainty $U(D)$ exceeds a predefined threshold, which we set empirically to 0.8. 
As shown in Fig.~\ref{fig:multi-reg}, for each pixel $p$ that belongs to $P_U$, we compute the intersection point $s$ between its primary ray $r$ and the surface. 
We then construct a spherical sampling space centered at $s$ with radius $H$. 
From this sphere, we uniformly sample $M$ points $m_i$ on its surface. 
For each sampled point, we generate an auxiliary ray $r_i$, which originates from $m_i$ and points toward $s$ with direction $f_i = \frac{s - m_i}{\|s - m_i\|}$. 
For each auxiliary ray $r_i$, we compute the rendered depth $\hat{D}_i^a$ using the volumetric rendering formula: 
\begin{equation}
\hat{D}_i^a = \sum_{p \in r_i} T_p \alpha_p t_p,
\end{equation}  
where $p$ denotes uniformly sampled points along $r_i$, with starting point $m_i$ and direction $f_i$. This value corresponds to the estimated distance from $m_i$ to the first surface intersection.  The multi-view consistency loss is then defined as:  
\begin{equation}
\mathcal{L}_{mv} = \sum_{p \in P_U} \frac{\sum_{i=1}^{M} v_i \, \big| H - \hat{D}_i^a \big|}{\sum_{i=1}^{M} v_i + \epsilon},
\end{equation}  
where $\epsilon$ is a small constant to avoid division by zero. The visibility indicator $v_i$ is defined as follows: if the first intersection point of the auxiliary ray $r_i$ coincides with that of the primary ray $r$, we set $v_i = 1$, and the theoretical depth $\hat{D}_i^a$ of $r_i$ should equal the sphere radius $H$. Otherwise, we set $v_i = 0$. This ensures that only mutually visible rays contribute to the consistency constraint, preventing invalid samples from affecting the loss.  

Although our multi-view consistency regularization and RayDF~\cite{RayDF} both involve consistency constraints, they are fundamentally distinct in their design philosophy, application scope, and implementation requirements. First, our constraint is specifically designed for neural SDF frameworks and is \textbf{selective, local}, activated only in high-uncertainty regions for targeted refinement. In contrast, RayDF is tailored for the ray--surface distance field representation and enforces a \textbf{global} consistency constraint uniformly across all rays. Second, our formulation is geometry-driven, deriving supervision from volumetric rendering in a local space, which contrasts with RayDF's classifier-driven approach. Finally, our method is lightweight and requires no auxiliary networks, unlike RayDF, which necessitates a separate classifier and a two-stage optimization, introducing overhead.

\begin{figure*}[t] 
\centerline{\includegraphics[width=1\linewidth]{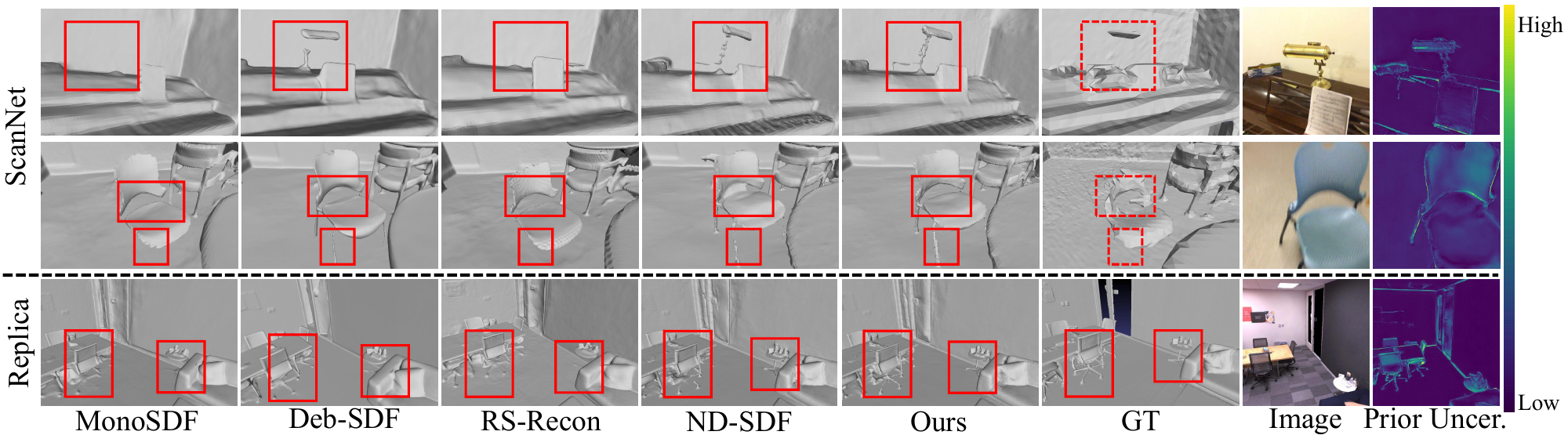}}
\vspace{-8pt}
\caption{Visualization results of reconstructed mesh on ScanNet and Replica, with details highlighted in red boxes. The rightmost column shows an example of the prior uncertainty for the corresponding region. Note that the ground-truth mesh of the ScanNet dataset is generated from RGB-D sensors using voxel hashing, and therefore it is not accurate in fine-detail regions. These structures are highlighted with red dashed boxes and can be verified from the RGB images. Our method achieves finer geometric structures compared to previous methods.}
\label{fig:performance}
\vspace{-17pt}
\end{figure*}

\textbf{Final objective.}
Finally, the overall loss function for GPU-SDF is as follows:
{\begin{equation}\label{equ:overall}
\mathcal{L} = \mathcal{L}_{recon}+\lambda_n^{kl}\mathcal{L}_n^{kl} +\lambda_d^{kl}\mathcal{L}_d^{kl}+\lambda_{e}\mathcal{L}_{e}+\lambda_{m}\mathcal{L}_{mv}.
\end{equation}}

\subsection{Implementation Details}\label{sec:imp}
During training, we randomly select four RGB images from the sequence together with their corresponding geometric priors. Using our model, we render the depth and normal maps and compute the normal deflection angle maps, as defined in ND-SDF\cite{NDSDF}. 
Based on the computed normal deflection angles, we apply importance sampling following \cite{DebSDF,NDSDF} to select 1,024 pixels from each image. The sampled spatial points are encoded using a layer-activated hash grid with resolutions ranging from $2^5$ to $2^{11}$ across 16 levels, with the initial activation set to 8. The encoded features $F_g$ are then input into the SDF, color, and edge decoders. Each decoder consists of fully connected layers with a size of $256 \times 256$. Rendering is performed as described in Sec.~\ref{sec:pre}, and the parameters of our neural SDF model are optimized by minimizing~\ref{equ:overall}. We use the Adam optimizer with a learning rate of $1\times10^{-3}$. The training runs for 128,000 iterations, taking approximately 24 GPU hours on a desktop PC with an NVIDIA RTX 4090. The weights of the loss function are set as: $\lambda_c = 1$, 
$\lambda_{eik} = 0.05$, 
$\lambda_{d} = 0.05$, 
$\lambda_{n} = 0.025$, 
$\lambda_n^{kl} = 0.025$, 
$\lambda_d^{kl} = 0.05$, 
$\lambda_{e} = 0.01$, and $\lambda_{m} = 512$. For multi-view consistency regularization, the number of points sampled on the sphere’s surface is set to $M = 1$, and the sphere radius is set to $H = 0.15$.

\section{Experiments}
\textbf{Datasets.}
We validate the effectiveness of GPU-SDF on 3 challenging indoor datasets: ScanNet\cite{scannet}, Replica\cite{replica}, ScanNet++\cite{scannet++}.
ScanNet is a large-scale 3D dataset of real-world indoor scenes.  
Replica is a high-quality, synthetically created indoor dataset. {
ScanNet++ is a high-fidelity real-world indoor dataset captured at high resolution.
}
For testing, we selected four representative scenes from ScanNet, eight from Replica, and three from ScanNet++.

\textbf{Metrics.}
We use the same evaluation metrics as previous work\cite{monosdf,DebSDF,NDSDF} to assess the meshes extracted from neural surface reconstruction: Accuracy, Completeness, Chamfer Distance, Precision, Recall, F-score and Normal Consistency.

\textbf{Baselines.}
{{
We compare GPU-SDF with four categories of methods: 
(1) Traditional MVS: COLMAP\cite{colmap}; 
(2) Neural implicit surface reconstruction from monocular RGB: VolSDF\cite{volsdf}, UNISURF\cite{unisurf}, Baked-Angelo\cite{bakedsdf}; 
(3) Neural implicit methods with additional priors : NeuRIS\cite{neuris}, MonoSDF\cite{monosdf}, RS-Recon\cite{RS-Recon}, NeRFPrior\cite{zhang2025nerfprior}, Deb-SDF\cite{DebSDF}, H2O-SDF\cite{H2O-SDF} , ND-SDF\cite{NDSDF}; 
(4) Gaussian Splatting–based methods: GSRec\cite{GSRec}. 
For fair comparison, we locally evaluate representative methods (MonoSDF, ND-SDF, GSRec) under the same environment and refer to reported results for the others.
}}

\subsection{Experiment Result}
The quantitative results on three datasets are reported in Table~\ref{tab:scan+rep} and  Table~\ref{tab:scannet++}. Our method, GPU-SDF, achieves state-of-the-art performance across these benchmarks. On global metrics, the numerical improvements are modest (e.g., an F-Score increase of 0.3 on ScanNet). This outcome is expected and highlights an inherent limitation of such metrics: they are dominated by large, low-frequency surfaces like walls and floors, which constitute the vast majority of the scene's surface area. Our approach, however, is specifically designed to excel at reconstructing complex, high-frequency details such as chair legs and railings. While these critical local improvements have a limited impact on the global score, their importance is undeniable for high-fidelity reconstruction.

Qualitative analysis provides a complement to global metrics. As shown in Fig.~\ref{fig:performance}, our method yields clearer reconstructions of intricate structures. For example, in regions with high geometry uncertainty, details such as chair legs---which are often missing or fragmented in other reconstructions---are recovered with higher clarity and completeness. This improved ability to capture geometric detail directly supports the core motivation of our work and demonstrates the effectiveness of our method in enhancing reconstruction fidelity where conventional approaches often struggle.

\begin{table}[t]
\centering
\vspace{-5pt}
\caption{Quantitative performance on ScanNet/Replica. Bold indicates the best performance.}
\label{tab:scan+rep}
\begin{tabular}{l l
                r@{/}l
                r@{/}l
                r@{/}l
                r@{/}l}
\toprule
Method & Prior 
       & \multicolumn{2}{c}{Acc$\downarrow$} 
       & \multicolumn{2}{c}{Comp$\downarrow$} 
       & \multicolumn{2}{c}{Chamfer$\downarrow$} 
       & \multicolumn{2}{c}{F-score$\uparrow$} \\
\midrule
COLMAP      & $\times$ & 4.7 & 3.0 & 23.5 & 9.5 & 14.1   & 6.3 & 53.7 & 65.8 \\
Unisurf      & $\times$ & 55.4 & 4.5 & 16.4 & 5.3 &  35.9   & 4.9 & 26.7 & 78.9 \\
VolSDF       & $\times$ & 41.4 & 4.4 & 12.0 & 8.3 & 26.7 & 7.2 & 34.6 & 69.5 \\
GSRec      & D+N      &  6.7 & 4.1 &  6.9 & 7.7 & 6.8 & 5.8 & 59.9 & 69.2 \\
NeuRIS       & N      & 5.0 & 3.4 & 4.9 & 7.0 & 5.0 & 5.2 & 69.2 & 75.4 \\
MonoSDF      & D+N      & 3.6 & 2.7 & 3.9 & 3.1 & 3.8 & 2.9 & 77.1 & 86.1 \\
RS-Recon     & D+N      & 4.0 & 2.7 & 4.0 & 2.5 & 3.6   & 2.6 & 79.4 & 91.7 \\ 
NeRFPrior$^\dagger$    & D+N      & 3.7 & --    & 4.2 & --    & --   & 3.8 & 78.2 & 81.3 \\ 
Deb-SDF       & D+N      & 3.6 & 2.8 & 4.0 & 3.0 & 3.8 & 2.9 & 78.5 & 88.3 \\
H2O-SDF$^\dagger$      & D+N      & 3.2 & --    & 3.7 & --    & 3.5 & --    & 79.9 & --   \\ 
ND-SDF       & D+N      & 3.1 & 2.7    & 3.6 & 2.8   & 3.4 & 2.6 & 82.0 & 91.6 \\
Ours         & D+N      & \textbf{3.1} & \textbf{2.6} 
             & \textbf{3.5} & \textbf{2.5}
             & \textbf{3.3} & \textbf{2.5}
             & \textbf{82.3}  & \textbf{92.4} \\
\bottomrule
\end{tabular}
\begin{tablenotes}
\footnotesize
\item[$\dagger$] Note: ``--'' denotes unavailable metrics. As of our evaluation, the source code for NeRFPrior and H2O-SDF was not publicly available, and their original papers did not report results on the ScanNet/Replica datasets.
\end{tablenotes}
\vspace{-18pt}
\end{table}

\begin{table}[t]
\centering
\caption{Quantitative Performance on ScanNet++. B.A. denotes Baked-Angelo. Bold indicates the best performance.}
\label{tab:scannet++}
\begin{tabularx}{\linewidth}{l l l*{4}{c}}
\hline
Methods & Priors & Metrics & Avg. & \makecell{0e75f\\3c4d9} & \makecell{036bc\\e3393} & \makecell{7f4d1\\73c9c} \\
\hline
\multirow{3}{*}{VolSDF} 
& \multirow{3}{*}{$\times$} & Acc $\downarrow$ & 6.8 & 5.3 & 5.3 & 9.7 \\
&  & Chamfer $\downarrow$ & 7.6 & 6.5 & 7.6 & 8.6 \\
&  & Prec $\uparrow$ & 44.2 & 39.5 & 47.5 & 45.7 \\
\hline

\multirow{3}{*}{B.A.} 
& \multirow{3}{*}{$\times$} & Acc $\downarrow$ & 23.6 & 12.1 & 16.8 & 42.0 \\
&  & Chamfer $\downarrow$ & 13.5 & 8.7 & 9.6 & 22.2 \\
&  & Prec $\uparrow$ & 50.1 & 52.5 & 52.0 & 45.7 \\
\hline

\multirow{3}{*}{MonoSDF} 
& \multirow{3}{*}{D+N} & Acc $\downarrow$ & 8.0 & 6.4 & 3.7 & 13.8 \\
&  & Chamfer $\downarrow$ & 5.4 & 4.6 & 3.7 & 8.0 \\
&  & Prec $\uparrow$ & 62.9 & 57.4 & 60.6 & 70.8 \\
\hline

\multirow{3}{*}{ND-SDF} 
& \multirow{3}{*}{D+N} & Acc $\downarrow$ & {6.8} & 5.4 & 3.7 & 11.2 \\
&  & Chamfer $\downarrow$ & {4.5} & 3.5 & \textbf{3.4} & 6.4 \\
&  & Prec $\uparrow$ & {67.7} & 66.2 & 64.8 & 72.1 \\
\hline

\multirow{3}{*}{Ours} 
& \multirow{3}{*}{D+N} & Acc $\downarrow$ & \textbf{6.1} & \textbf{4.2} & \textbf{3.7} & \textbf{10.3} \\
&  & Chamfer $\downarrow$ & \textbf{4.3} & \textbf{3.4} & 3.6 & \textbf{6.0} \\
&  & Prec $\uparrow$ & \textbf{68.9} & \textbf{67.9} & \textbf{64.8} & \textbf{73.9} \\
\hline
\end{tabularx}
\vspace{-5pt}
\end{table}

\begin{table}[t]
\centering
\caption{Ablation Studies Results for key components of GPU-SDF.Bold indicates the best performance.}
\label{tab:ab++}
    \begin{tabularx}{0.9\linewidth}{ c *{3}{>{\centering\arraybackslash}X}}
\toprule
Method & Acc. $\downarrow$ & Chamfer. $\downarrow$ & Prec.$\uparrow$ \\
\midrule
ND-SDF                & 11.2 & 6.4 & 72.1 \\
\hspace{5pt}+D.U.(Horiz.) & 10.8 & 6.3 & 73.1 \\
\hspace{5pt}+D.U.(Vert.)  & 10.6 & 6.2 & 72.8 \\
\hspace{5pt}+D.U.(Full)   & 10.7 & 6.2 & 73.3 \\
\hspace{5pt}+D.U.+N.U.    & 10.4 & 6.0 & 73.7 \\
\midrule
w/o EDF              & 11.0 & 6.4 & 72.7\\
w/o MC               & 10.8 & 6.3 & 72.7 \\
Full Model           &\textbf{10.3} & \textbf{6.0} & \textbf{73.9}   \\
\midrule
MonoSDF              &13.8 & 8.0 & 70.8   \\
MonoSDF+Ours    & \textbf{11.4} & \textbf{6.9} & \textbf{71.6}   \\
\bottomrule
\end{tabularx}
\vspace{-15pt}
\end{table}

\subsection{Ablation Studies}\label{sec:ab}
We conducted a series of ablation experiments on the ScanNet++ dataset to evaluate the contributions of each key component in our framework. Quantitative results are summarized in Table~\ref{tab:ab++}.

\textbf{Effectiveness of Prior Uncertainty.} 
We first analyze the design of the prior uncertainty estimation module, using ND-SDF as the baseline and incrementally adding different strategies. We denote depth uncertainty as \emph{D.U.} and normal uncertainty as \emph{N.U.} The results show that even when using only depth uncertainty derived from a single type of data augmentation (\emph{+D.U.(Horiz.)} or \emph{+D.U.(Vert.)}), reconstruction accuracy improves. Combining both horizontal and vertical flips (\emph{+D.U.(Full)}) produces more reliable uncertainty estimates, as illustrated in Fig.~\ref{fig:uncer}, and further enhances reconstruction quality. The largest gain is achieved by combining both depth and normal uncertainty (\emph{+D.U.+N.U.}), which reduces the Chamfer distance from the baseline value of 0.064 to 0.060 and increases precision from 72.1 to 73.7. These results highlight that our uncertainty modeling is an essential component for guiding the reconstruction process.

\begin{figure}[t] 
\centerline{\includegraphics[width=1\linewidth]{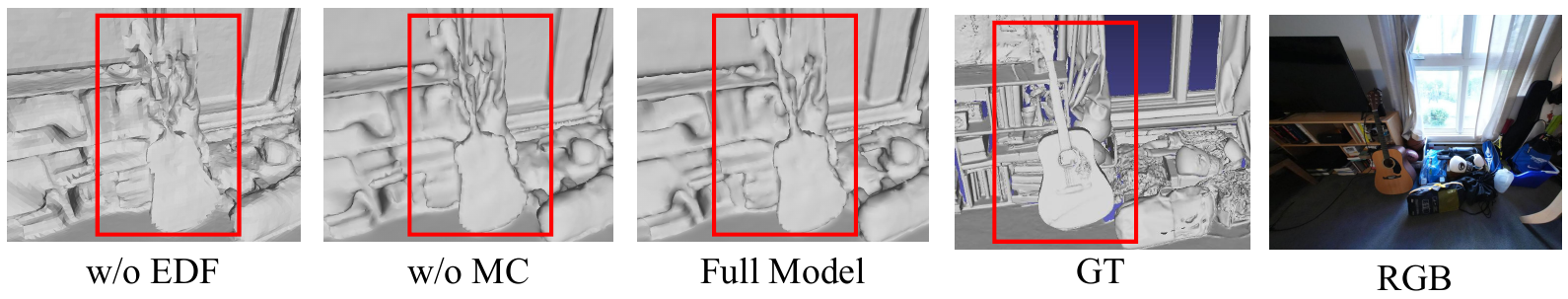}}
\vspace{-10pt}
\caption{
{{Visualization results of ablation studies, with details highlighted in red boxes. Full Model achieves the finest detail in thin structures, demonstrating the effectiveness of our method.}}
}\label{fig:Ablation}
\vspace{-18pt}
\end{figure}

\textbf{Impact of Additional Geometric Constraints.}  
We evaluate the necessity of the two geometric constraints introduced for high-uncertainty regions: the edge distance field (EDF) and multi-view consistency (MC). We conduct two controlled comparisons, removing EDF (\emph{w/o EDF}) and removing MC (\emph{w/o MC}), against full model. As shown in Table~\ref{tab:ab++}, removing either constraint causes a performance drop, confirming that both are essential for suppressing artifacts in under-constrained regions and preserving structural integrity. The visualization results are shown in Fig.~\ref{fig:Ablation}.

\textbf{Generalization as a Plug-in Module.} 
Finally, we test our framework as a plug-in that integrates seamlessly into other neural SDF pipelines to enhance their performance. To this end, we apply our method to another mainstream baseline, MonoSDF \cite{monosdf}. The results are compelling: after integration (\emph{MonoSDF+Ours}), all metrics show substantial improvements. In particular, the Chamfer distance decreases from 0.080 to 0.069, and Precision increases from 70.8 to 71.6. This experiment demonstrates the strong generalization ability of our approach, indicating that it is not tailored to a specific architecture but serves as a versatile module for improving neural surface reconstruction quality.

\section{Conclusion And Discussion}
In this paper, we introduce GPU-SDF, a novel neural implicit indoor surface reconstruction method. Our core contributions are twofold. First, we introduce a self-supervised uncertainty estimation with guided supervision, allowing reliable use of geometric priors while preserving information in high-uncertainty regions. Second, to address the resulting under-constrained optimization, we further incorporate an edge distance field and multi-view consistency regularization, which enhance thin structures and fine details. Experiments on challenging benchmarks verified the effectiveness of GPU-SDF and its versatility as a plug-in to existing SDF methods, enabling more robust and reliable 3D reconstruction with imperfect priors. However, the regions corresponding to unseen viewpoints remain unconstrained, and future work will focus on enhancing the reconstruction quality of regions that are occluded or not visible to the camera.

\bibliographystyle{IEEEtran}
\bibliography{IEEEexample.bib}

\end{document}